%% file: gen.tex
\documentclass[twocolumn,10pt]{article}

\usepackage{acnn98}
\usepackage{times}
\usepackage{amsmath}
\usepackage{amsthm}
\usepackage{amssymb}
\usepackage{epsf}
\usepackage{epic}
\usepackage{eepic}
\usepackage{ecltree}

\input{mathdefs}

\begin{document}

\input{topmatter}

\maketitle

\input{abstract}

\input{guts}

\bibliographystyle{plain}
\bibliography{bib}

\end{document}

%% file: mathdefs.tex
\renewcommand{\glossary}[2]{}
\def\addcontentsline#1#2#3{}

\newcommand{\be}{\begin{equation*}}

\renewcommand{\[}{\left[}
\renewcommand{\]}{\right]}

\renewcommand{\hbar}{{\overline h}}

\newcommand{\Jt}{{\tilde{J}}}
\newcommand{\Js}{{J^*}}

\newcommand{\R}{{\mathbb R}}

\newcommand{\argmin}{{\mbox{argmin}}}

%% file: topmatter.tex
\title{TDLeaf($\lambda$): Combining Temporal Difference Learning with Game-Tree Search.}

\author{
~~~~~~~~~~~~~~Jonathan Baxter~~~~~~~~~~~~~~~~~~~~~~~~~~~~~~~~~~~~~~Andrew Tridgell \& Lex Weaver\\
Department of Systems Engineering~~~~~~~~~~~~~~~~~~~~Department of Computer Science\\
~~Australian National University~~~~~~~~~~~~~~~~~~~~~~~~~~Australian National University\\
~~~Canberra 0200, Australia~~~~~~~~~~~~~~~~~~~~~~~~~~~~~~~~~~~Canberra 0200, Australia\\
{\tt \{Jon.Baxter,Andrew.Tridgell,Lex.Weaver\}@anu.edu.au}}

%% file: abstract.tex
\begin{abstract}
In this paper we present TDLeaf($\lambda$), a variation on the
TD($\lambda$) algorithm that enables it to be used in conjunction with
minimax search. We present some experiments in both chess and backgammon 
which demonstrate its utility and provide comparisons with TD($\lambda$) and 
another less radical variant, TD-directed($\lambda$). In particular, our 
chess program, ``KnightCap,'' used TDLeaf($\lambda$) to learn its
evaluation function while playing on the Free Internet Chess Server
(FICS, {\tt fics.onenet.net}). It improved from a 1650 rating to a 2100 
rating in just 308 games.  We discuss some of the reasons for this
success and the relationship between our results and
Tesauro's results in backgammon.
\end{abstract}

%% file: guts.tex
\section{Introduction}
TD($\lambda$), developed by
Sutton \cite{sutton88}, has its roots in the learning algorithm of Samuel's checkers program \cite{samuel59}. It is an elegant algorithm for approximating the
expected long term future cost of a stochastic
dynamical system as a function of the current state. The mapping from
states to future cost is implemented by a parameterised function
approximator such as a neural network. The parameters are updated
online after each state transition, or in batch updates
after several state transitions. The goal of the algorithm is to
improve the cost estimates as the number of observed state transitions
and associated costs increases.

Tesauro's TD-Gammon is perhaps the most remarkable success of TD($\lambda$). It is a neural network backgammon player that has proven itself to be
competitive with the best human backgammon players \cite{Tesauro94}.

Many authors have discussed the peculiarities of backgammon that make
it particularly suitable for Temporal Difference learning with
self-play \cite{tesauro92,schraudolph94,pollack96}. Principle among
these are {\em speed of play}: TD-Gammon learnt from several hundred
thousand games of self-play, {\em representation smoothness}: the
evaluation of a backgammon position is a reasonably smooth function of
the position (viewed, say, as a vector of piece counts), making it
easier to find a good neural network approximation, and {\em
stochasticity}: backgammon is a random game which forces at least a
minimal amount of exploration of search space.

As TD-Gammon in its original form only searched one-ply ahead,
we feel this list should be appended with: {\em shallow search is good
enough against humans}. There are two possible reasons for this;
either one does not gain a lot by searching deeper in backgammon
(questionable given that recent versions of TD-Gammon search to three-ply for a significant performance improvement), or humans are incapable 
of searching deeply and so TD-Gammon is only competing
in a pool of shallow searchers.

In contrast, finding a representation for chess, othello or Go which
allows a small neural network to order moves at one-ply with near
human performance is a far more difficult task \cite{thrun95b,walker93,schraudolph94}. For
these games, reliable tactical evaluation is difficult to achieve
without deep search. This requires an exponential increase
in the number of positions evaluated as the search depth increases.
Consequently, the computational cost of the evaluation function has to be low
and hence, most chess and
othello programs use linear functions.

In the next section we look at {\em reinforcement learning} (the broad category into which TD($\lambda$) falls), and then in subsequent sections we look at TD($\lambda$) in some detail and introduce two variations on the theme: TD-directed($\lambda$) and TDLeaf($\lambda$). The first uses minimax search to generate better training data, and the second, TDLeaf($\lambda$), is used to learn an evaluation function for use in deep minimax search.

\section{Reinforcement Learning}
The popularly known and best understood learning techniques fall into the category of {\em supervised learning}. This category is distinguished  by the fact that for each input upon which the system is trained, the ``correct'' output is known. This allows us to measure the error and use it to train the system.

For example, if our system maps input $X_{i}$ to output $Y'_{i}$, then, with $Y_{i}$ as the ``correct'' output, we can use $(Y'_{i} - Y_{i})^{2}$ as a measure of the error corresponding to $X_{i}$. Summing this value across a set of training examples yields an error measure of the form $\sum_{i}(Y'_{i}\nolinebreak-\nolinebreak Y_{i})^{2}$, which can be used by training techniques such as back propagation.

Reinforcement learning differs substantially from supervised learning in that the ``correct'' output is not known. Hence, there is no direct measure of error, instead a scalar reward is given for the responses to a series of inputs.

Consider an {\em agent} reacting to its {\em environment} (a generalisation of the two-player game scenario). Let $S$ denote the set of all possible environment states. Time proceeds with the agent performing actions at discrete time steps $t=1, 2, ...$ . At time $t$ the agent finds the environment in state $x_{t}\in S$, and has available a set of actions $A_{x_{t}}$. The agent chooses an action $a_{t}\in A_{x_{t}}$, which takes the environment to state $x_{t+1}$ with probability $p(x_{t},x_{t+1},a_{t})$. After a determined series of actions in the environment, perhaps when a goal has been achieved or has become impossible, the scalar reward, $r(x_{N})$ where $N$ is the number of actions in the series, is awarded to the agent. These rewards are often discrete, eg: ``1'' for success, ``-1'' for failure, and ``0'' otherwise.

For ease of notation we will assume all series of actions have a fixed length of
$N$ (this is not essential). If we assume that the agent chooses its
actions according to some function $a(x)$ of the current state $x$ (so
that $a(x) \in A_{x}$), the expected reward from each state $x\in S$ is given
by 
\begin{equation}
\label{expreward}
\Js(x) := E_{x_N|x} r(x_N),
\end{equation}
where the expectation is with respect to the transition probabilities
$p(x_t,x_{t+1}, a(x_t))$. 

Once we have $\Js(u)$, we can ensure that actions are chosen optimally in any state by using the following equation to minimise the expected reward for the environment ie: the other player in the game.

\begin{equation}
\label{optimalactioneq}
a^*(x) := \argmin_{a\in A_x} \Js(x'_a, w).
\end{equation}

For very large state spaces $S$ it is not possible store the value of
$\Js(x)$ for every $x\in S$, so instead we might try to approximate
$J^*$ using a parameterised function class $\Jt\colon S\times \R^k \to
\R$, for example linear function, splines, neural networks,
etc. $\Jt(\cdot,w)$ is assumed to be a differentiable function of its
parameters $w=(w_1,\dots,w_k)$. The aim is to find $w$ so that $\Jt(x,w)$ is ``close to'' $\Js(u)$, at least in so far as it generates the correct ordering of moves.

This approach to learning is quite different from that of supervised learning where the aim is to minimise an explicit error measurement for each data point.

Another significant difference between the two paradigms is the nature of the data used in training. With supervised learning it is fixed, whilst with reinforcement learning the states which occur during training are dependent upon the agent's choice of action, and thus on the training algorithm which is modifying the agent. This dependency complicates the task of proving convergence for TD$(\lambda)$ in the general case \cite{Bertsekas96}.

\section{The TD($\lambda$) algorithm}
Temporal Difference learning or TD$(\lambda)$, is perhaps the best known of the reinforcement learning algorithms. It provides a way of using the scalar rewards such that existing supervised training techniques can be used to tune the function approximator. Tesauro's TD-Gammon for example, uses back propagation to train a neural network function approximator, with TD$(\lambda)$ managing this process and calculating the necessary error values.

Here we consider how TD$(\lambda)$ would be used to train an agent playing a two-player game, such as chess or backgammon.

Suppose $x_1,\dots,x_{N-1}, x_N$ is a sequence of states in one game.
For a given parameter vector $w$, define the {\em temporal difference}
associated with the transition $x_t\rightarrow x_{t+1}$ by
\begin{equation}
\label{tempdiff}
d_t := \Jt(x_{t+1},w) - \Jt(x_t,w).
\end{equation}
Note that $d_t$ measures the difference between the reward predicted
by $\Jt(\cdot,w)$ at time $t+1$, and the reward predicted by
$\Jt(\cdot,w)$ at time $t$. The true evaluation function $\Js$ has the
property
$$
E_{x_{t+1}|x_t} \[\Js(x_{t+1}) - \Js(x_t)\] = 0,
$$
so if $\Jt(\cdot,w)$ is a good approximation to $\Js$, 
$E_{x_{t+1}|x_t} d_t$ should be close to zero. 
For ease of notation we will assume that $\Jt(x_N,w) = r(x_N)$ always,
so that the final temporal difference satisfies
$$
d_{N-1} = \Jt(x_{N},w) - \Jt(x_{N-1},w) =  r(x_N) -  \Jt(x_{N-1},w).
$$
That is, $d_{N-1}$ is the difference between the true outcome of the
game and the prediction at the penultimate move.  

At the end of the game, the TD($\lambda$) algorithm updates the
parameter vector $w$ according to the formula
\begin{equation}
\label{tdeq}
w := w + \alpha\sum_{t=1}^{N-1} \nabla\Jt(x_t,w)\[\sum_{j=t}^{N-1} 
\lambda^{j-t}d_t\]
\end{equation}
where $\nabla\Jt(\cdot,w)$ is the vector of partial derivatives of
$\Jt$ with respect to its parameters. The positive parameter $\alpha$
controls the learning rate and would typically be ``annealed'' towards
zero during the course of a long series of games. The parameter
$\lambda\in [0,1]$ controls the extent to which temporal differences
propagate backwards in time. To see this, compare equation
\eqref{tdeq} for $\lambda=0$:
\begin{align}
\label{td0}
w := &w + \alpha\sum_{t=1}^{N-1} \nabla\Jt(x_t,w)d_t \notag\\
   = &w + \alpha\sum_{t=1}^{N-1} \nabla\Jt(x_t,w)
\[\Jt(x_{t+1},w) - \Jt(x_t,w)\]
\end{align}
and $\lambda=1$:
\begin{equation}
\label{td1}
w := w + \alpha\sum_{t=1}^{N-1} \nabla\Jt(x_t,w)
\[r(x_N) - \Jt(x_t,w)\].
\end{equation}
Consider each term contributing to the sums in equations \eqref{td0}
and \eqref{td1}.  For $\lambda=0$ the parameter vector is being
adjusted in such a way as to move $\Jt(x_t,w)$ --- the predicted reward
at time $t$ --- closer to $\Jt(x_{t+1},w)$ --- the predicted reward at
time $t+1$.  In contrast, TD(1) adjusts the parameter vector in such
away as to move the predicted reward at time step $t$ closer to the
final reward at time step $N$. Values of $\lambda$ between zero and one
interpolate between these two behaviours. Note that \eqref{td1} is
equivalent to gradient descent on the error function $E(w) :=
\sum_{t=1}^{N-1}\[r(x_N) - \Jt(x_t,w)\]^2$. 

Tesauro \cite{tesauro92,Tesauro94} and those who have replicated his work with backgammon, report that the results are insensitive to the value of $\lambda$ and commonly use a value around 0.7. Recent work by Beale and Smith \cite{beal97} however, suggests that in the domain of chess there is greater sensitivity to the value of $\lambda$, with it perhaps being profitable to dynamically tune $\lambda$. 

Successive parameter updates according to the TD($\lambda$) algorithm
should, over time, lead to improved predictions of the expected reward
$\Jt(\cdot, w)$. Provided the actions $a(x_t)$ are independent of the
parameter vector $w$, it can be shown that for {\em linear}
$\Jt(\cdot,w)$, the TD($\lambda$) algorithm converges to a
near-optimal parameter vector \cite{Tsitsikilis97}. Unfortunately,
there is no such guarantee if $\Jt(\cdot,w)$ is non-linear
\cite{Tsitsikilis97}, or if $a(x_t)$ depends on $w$
\cite{Bertsekas96}.

\section{Two New Variants}
\begin{figure*}
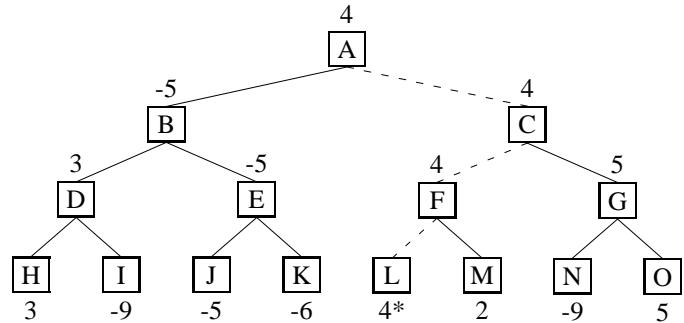

\begin{center}
\begin{bundle}{\shortstack{4 \\ \framebox[5mm]{A}}}
\offinterlineskip
\drawwith{\drawwith{\dashline{3}}\drawline}	
\chunk[-5]{
	\begin{bundle}{\framebox[5mm]{B}}
	\drawwith{\drawline}
	\chunk[3]{
		\begin{bundle}{\framebox[5mm]{D}}
		\drawwith{\drawline}
		\chunk{\shortstack{\framebox[5mm]{H} \\ 3}}
		\chunk{\shortstack{\framebox[5mm]{I} \\ -9}}
		\end{bundle}}
	\chunk[-5]{
		\begin{bundle}{\framebox[5mm]{E}}
		\chunk{\shortstack{\framebox[5mm]{J} \\ -5}}
		\chunk{\shortstack{\framebox[5mm]{K} \\ -6}}
		\end{bundle}}
	\end{bundle}}
\chunk[4]{
	\begin{bundle}{\framebox[5mm]{C}}
	\drawwith{\drawwith{\drawline}\dashline{3}}
	\chunk[4]{
		\begin{bundle}{\framebox[5mm]{F}}
		\chunk{\shortstack{\framebox[5mm]{L} \\ 4*}}
		\chunk{\shortstack{\framebox[5mm]{M} \\ 2}}
		\end{bundle}}
	\chunk[5]{
		\begin{bundle}{\framebox[5mm]{G}}
		\drawwith{\drawline}
		\chunk{\shortstack{\framebox[5mm]{N} \\ -9}}
		\chunk{\shortstack{\framebox[5mm]{O} \\ 5}}
		\end{bundle}}
	\end{bundle}}
\end{bundle}
\end{center}
\caption{Full breadth, 3-ply search tree illustrating the minimax rule
for propagating values. Each of the leaf nodes (H--O) is given a score
by the evaluation function, $\Jt(\cdot, w)$. These scores are then
propagated back up the tree by assigning to each opponent's internal
node the minimum of its children's values, and to each of our internal
nodes the maximum of its children's values. The principle variation is
then the sequence of best moves for either side starting from the root
node, and this is illustrated by a dashed line in the figure. Note
that the score at the root node A is the evaluation of the leaf node
(L) of the principal variation. As there are no ties between any
siblings, the derivative of A's score with respect to the parameters $w$
is just $\nabla\Jt(L, w)$.
\label{minimaxfig}}
\end{figure*}

For argument's sake, assume any action $a$ taken in state $x$ leads to
predetermined state which we will denote by $x'_a$. Once an
approximation $\Jt(\cdot,w)$ to $\Js$ has been found, we can use it to
choose actions in state $x$ by picking the action $a\in A_x$ whose
successor state $x'_a$ minimizes the opponent's expected
reward\footnote{If successor states are only determined stochastically
by the choice of $a$, we would choose the action minimizing the
expected reward over the choice of successor states.}:
\begin{equation}
\label{actioneq}
\tilde{a}(x) := \argmin_{a\in A_x} \Jt(x'_a, w).
\end{equation}
This was the strategy used in TD-Gammon. Unfortunately, for games like
othello and chess it is difficult to accurately evaluate a
position by looking only one move or {\em ply} ahead. Most programs
for these games employ some form of {\em minimax} search. In minimax
search, one builds a tree from position $x$ by examining all possible
moves for the computer in that position, then all possible moves for
the opponent, and then all possible moves for the computer and so on
to some predetermined depth $d$. The leaf nodes of the tree are then
evaluated using a heuristic evaluation function (such as
$\Jt(\cdot,w)$), and the resulting scores are propagated back up the
tree by choosing at each stage the move which leads to the best
position for the player on the move.  See figure \ref{minimaxfig} for
an example game tree and its minimax evaluation. With reference to the
figure, note that the evaluation assigned to the root node is the
evaluation of the leaf node of the {\em principal variation}; the
sequence of moves taken from the root to the leaf if each side chooses
the best available move.

\setlength{\GapWidth}{7mm}

Our TD-directed($\lambda$) variant utilises minimax search by allowing play to be guided by minimax, but still defines the temporal differences to be the differences in the evaluations of successive board positions occurring during the game, as per 
equation (\ref{tempdiff}).

Let $\Jt_d(x,w)$ denote the evaluation obtained for state $x$ by
applying $\Jt(\cdot,w)$ to the leaf nodes of a depth $d$ minimax
search from $x$. Our aim is to find a parameter vector $w$ such that
$\Jt_d(\cdot,w)$ is a good approximation to the expected reward
$\Js$. One way to achieve this is to apply the TD($\lambda$)
algorithm to $\Jt_d(x,w)$. That is, for each sequence of positions
$x_1,\dots,x_N$ in a game we define the temporal differences
\begin{equation}
\label{tempdiffd}
d_t := \Jt_d(x_{t+1},w) - \Jt_d(x_t,w)
\end{equation}
as per equation \eqref{tempdiff}, and then the TD($\lambda$) algorithm
\eqref{tdeq} for updating the parameter vector $w$ becomes
\begin{equation}
\label{tdeqd}
w := w + \alpha\sum_{t=1}^{N-1} \nabla\Jt_d(x_t,w)\[\sum_{j=t}^{N-1} 
\lambda^{j-t}d_t\].
\end{equation}
One problem with equation \eqref{tdeqd} is that for $d>1$,
$\Jt_d(x,w)$ is not a necessarily a differentiable function of $w$ for
all values of $w$, even if $\Jt(\cdot,w)$ is everywhere
differentiable. This is because for some values of $w$ there will be
``ties'' in the minimax search, i.e.  there will be more than one best
move available in some of the positions along the principal variation,
which means that the principal variation will not be unique. Thus, the evaluation assigned to the root node,
$\Jt_d(x,w)$, will be the evaluation of any one of a number of leaf
nodes.

Fortunately, under some mild technical assumptions on the behaviour of
$\Jt(x,w)$, it can be shown that for all states $x$ and
for ``almost all'' $w\in \R^k$, $\Jt_d(x,w)$ is a differentiable
function of $w$.  Note that $\Jt_d(x,w)$ is also a continuous function
of $w$ whenever $\Jt(x,w)$ is a continuous function of $w$. This
implies that even for the ``bad'' pairs $(x,w)$, $\nabla\Jt_d(x,w)$ is
only undefined because it is multi-valued. Thus we can still
arbitrarily choose a particular value for $\nabla\Jt_d(x,w)$ if $w$
happens to land on one of the bad points.

Based on these observations we modified the TD($\lambda$) algorithm to
take account of minimax search: instead of
working with the root positions $x_1,\dots,x_N$, the TD($\lambda$)
algorithm is applied to the leaf positions found by minimax search
from the root positions. We call this algorithm
TDLeaf($\lambda$).

\section{Experiments with Chess}
\label{expsec}
In this section we describe several experiments in
which the TDLeaf($\lambda$) and TD-directed($\lambda$) algorithms were used to train the weights of
a linear evaluation function for our chess program, called KnightCap.

For our main experiment we took KnightCap's evaluation function and set
all but the material parameters to zero. The material parameters were
initialised to the standard ``computer'' values\footnote{1 for a pawn, 4 for a
knight, 4 for a bishop, 6 for a rook and 12 for a queen.}.  With these
parameter settings KnightCap was
started on the Free Internet Chess server (FICS, {\tt
fics.onenet.net}). To establish its rating, 25 games were played 
without modifying the
evaluation function, after which it had a blitz (fast time control) rating of
$1650\pm50$\footnote{After some experimentation, we have estimated the standard deviation of FICS ratings to be 50 ratings points.}. We then turned on
the TDLeaf($\lambda$) learning algorithm, with $\lambda=0.7$ and the
learning rate $\alpha=1.0$. The value of $\lambda$ was chosen arbitrarily, while $\alpha$ was set high enough to ensure rapid
modification of the parameters. 

After only 308 games, KnightCap's rating climbed to $2110\pm50$.
This rating puts KnightCap at
the level of US Master. 

We repeated the experiment using TD-directed($\lambda$), and observed a 200 point rating rise over 300 games. A significant improvement, but slower than TDLeaf($\lambda$).

There are a number of reasons for KnightCap's remarkable rate of improvement. 
\begin{enumerate}
\item 
\label{p2}
KnightCap started out with
intelligent material parameters. This put it close in parameter space
to many far superior parameter settings. 
\item 
\label{p3}
Most players on FICS prefer to play opponents of similar strength, and
so KnightCap's opponents improved as it did. Hence it received both positive and negative feedback from its
games.
\item 
\label{p5} 
KnightCap was not learning by self-play.
\end{enumerate}

To investigate the importance of some of these reasons, we
conducted several more experiments. \\

\noindent {\em Good initial conditions.} \\ A second
experiment was run in which KnightCap's coefficients were all
initialised to the value of a pawn.  

Playing with this initial weight setting KnightCap had a blitz rating
of $1260\pm50$.  After more than 1000 games on FICS KnightCap's rating
has improved to about $1540\pm50$, a 280 point gain. This is a much slower
improvement than the original experiment, and makes it clear that starting 
near a
good set of weights is important for fast convergence. \\

\noindent {\em Self-Play} \\ Learning by self-play was extremely
effective for TD-Gammon, but a significant reason for this is the
stochasticity of backgammon.
However, chess is a deterministic game and self-play by a
deterministic algorithm tends to result in a large number of
substantially similar games. This is not a problem if the games seen
in self-play are ``representative'' of the games played in practice,
however
KnightCap's self-play games with only non-zero material weights are
very different to the kind of games humans of the same level would
play.

To demonstrate that learning by self-play for KnightCap is not as
effective as learning against real opponents, we ran another
experiment in which all but the material parameters were initialised
to zero again, but this time KnightCap learnt by playing against
itself. After 600 games (twice as many as in the original FICS
experiment), we played the resulting version against the good version that learnt on FICS, in a 100 game match with the weight values fixed. The FICS trained  version won 89 points to the self-play version's 11.

\input{backgammon}

\section{Discussion and Conclusion}
\label{conclusion}
We have introduced TDLeaf($\lambda$), a variant of TD($\lambda$)
for training an evaluation function used in minimax
search. The only extra requirement of the algorithm is that the
leaf-nodes of the principal variations be stored throughout the game. 

We presented some experiments in which a chess evaluation function was
trained by on-line play against a mixture of human and computer
opponents. The experiments show both the importance of ``on-line''
sampling (as opposed to self-play), and the need to start near a good
solution for fast convergence. 

We compared training using leaf nodes (TDLeaf($\lambda$)) with
training using root nodes, both in chess with a linear evaluation
function and 5-10 ply search, and in backgammon with a
one hidden layer neural-network evaluation function and 2-ply search. 
We found a significant improvement training on the leaf nodes in
chess, which can be attributed to the substantially different
distribution over leaf nodes compared to root nodes. No such
improvement was observed for backgammon which suggests that the
optimal network to use in 1-ply search is close to the optimal network
for 2-ply search. 

On the theoretical side, it has recently been shown that TD($\lambda$)
converges for linear evaluation functions \cite{Tsitsikilis97}. An
interesting avenue for further investigation would be to determine
whether TDLeaf($\lambda$) has similar convergence properties.

%% file: backgammon.tex
\section{Backgammon Experiment}
\label{bgsec}

For our backgammon experiment we were fortunate to have Mark Land (University of California, San Diego) provide us with the source code for his LGammon program which has been implemented along the lines of Tesauro's TD-Gammon\cite{tesauro92,Tesauro94}.

Along with the code for LGammon, Land also provided a set of weights for the neural network. The weights were used by LGammon when playing on the First Internet Backgammon Server (FIBS, fibs.com), where LGammon achieved a rating which ranged from 1600 to 1680, significantly above the mean rating across all players of about 1500. For convenience, we refer to the weights as the {\em FIBS weights}.  

Using LGammon and the FIBS weights to directly compare searching to two-ply against searching to one-ply, we observed that two-ply is stronger by 0.25 points-per-game, a significant difference in backgammon. Further analysis showed that in 24\% of positions, the move recommended by a two-ply search differed from that recommended by a one-ply search.

Subsequently, we decided to investigate how well TD-directed($\lambda$) and TDLeaf($\lambda$), both of which can search more deeply, might perform.
Our experiment sought to determine whether either TD-directed($\lambda$) or TDLeaf($\lambda$) could find better weights than standard TD($\lambda$).

To test this, we suitably modified the algorithms to account for the stochasticity inherent in the game, and took two copies of the FIBS weights --- the end product of a standard TD($\lambda$) training run of 270,000 games. We trained one copy using TD-directed($\lambda$) and the other using TDLeaf($\lambda$). Each network was trained for 50000 games and then played against the unmodified FIBS weights for 1600 games, with both sides searching to two-ply and the match score recorded.

The results fluctuated around parity with the FIBS weights (the product of TD($\lambda$) training), with no statistically significant change in performance being observed. This suggests that the solution found by TD($\lambda$), is either at or near the optimal for two-ply play.